AI.f(T)

# From Control Boundary to Insurance Claim:

*Reconstructing AI-Mediated Losses Through the CER Framework*

Alex Leung, Rex Zhang, Kentaroh Toyoda, SiewMei Loh
AIFT

*June 2, 2026*

**ABSTRACT**

AI losses that arise through an insured organization's generative or agentic AI system require state reconstruction, not merely event reconstruction, because the relevant state changes as the system reasons, retrieves, calls tools, and acts. The relevant question is not only what loss occurred, but what the system was allowed to do, what it actually did, and whether that reconstructed loss can support insurance claim recovery. This paper addresses losses in which the insured's AI system is in the causal chain, including externally triggered failures such as prompt injection, retrieval-augmented generation (RAG) poisoning, malicious tool output, credential misuse, and data poisoning.

Specifically, this paper introduces CER, a use-case-level diagnostic for AI residual risk transfer. C (control boundary) asks whether the system had an enforceable operating envelope. E (evidence reconstruction) asks whether the system state and causal chain can be reconstructed from retained artifacts. R (insurance response) asks whether the reconstructed loss is insured: whether insurance coverage is available in the market and placed for the insured, together with the proof needed to support insurance claim recovery.

The paper makes three contributions: it defines the AI-specific reconstruction problem, operationalizes that problem through CER, and specifies claim-grade evidence for AI reconstruction. Public examples include the reported PocketOS and Replit agentic database-deletion incidents and Moffatt v. Air Canada as an adjudicated output/reliance case.

## I. INTRODUCTION

Organizations increasingly deploy generative and agentic AI systems that retrieve data, call tools, and take actions on their behalf. When one of these systems contributes to a loss, three questions immediately decide whether the loss can be transferred to an insurer: what the system was permitted to do, what it actually did, and whether the resulting loss can be proven and matched to coverage. Current AI-risk frameworks and insurance materials address parts of this but not the use-case-level proof chain that links an AI system's state to an insured loss.

The motivating example is an AI agent exercising real operational authority. Public reports describe a 2026 PocketOS incident in which a Cursor coding agent, reportedly running Claude Opus 4.6, deleted a production database and backups after finding and using a broadly scoped infrastructure API token. The central point is simple: the agent was reportedly instructed not to touch production, but the production boundary was not technically enforceable.

**Related work in brief.** AI-risk frameworks such as the OWASP Top 10 for LLMs and OWASP Top 10 for Agentic Applications [1–3] and the NIST AI Risk Management Framework [5] catalog hazards and controls; insurance studies document coverage demand, silent-AI exposure, and emerging exclusions [13–17]; and the AI-accountability and auditing literature identifies gaps in proving how deployed systems behaved [27, 28]. These works are largely either control-facing or coverage-facing. None specifies the post-loss, use-case-level proof chain that connects an insured AI system's state to a recognized loss and then to an insurance response. Section III discusses this literature in more detail.

**Problem.** For a loss arising through an insured's AI system, current practice cannot reliably establish three things at once: that the system had an enforceable operating boundary, that its actual state and causal chain can be reconstructed from retained evidence, and that the reconstructed loss maps to available, placed insurance coverage. When any of the three is missing, residual AI risk stays with the organization, often without it realizing.

**Proposal.** We propose CER, a use-case-level diagnostic that scores these three links explicitly. C (control boundary) asks whether the AI system had an enforceable operating envelope; E (evidence reconstruction) asks whether the system state and causal chain can be reconstructed from retained artifacts to a claim-grade standard; and R (insurance response) asks whether the reconstructed loss is insured: whether suitable coverage is available in the market and placed for the insured. CER scores each link on a 0–3 scale, combines the scores into a diagnostic state, and treats a loss as transfer-aligned only when all three links are present and connected.

**Scope boundary.** CER applies when the insured organization's AI system materially contributes to the loss. The trigger may be internal or external. Prompt injection, RAG poisoning, malicious tool output, credential misuse, and data poisoning are in scope when they act through the insured's AI system. A purely external AI-enabled attack, such as third-party deepfake fraud that deceives an employee without involving the insured's AI system, may still trigger other insurance covers [22], but it is not scored under CER.

**Boundary rule for edge cases.** The scenario is in scope when the insured's AI system processes the malicious input and the loss flows from the AI's resulting output or action. It is out of scope when AI appears only in the attacker's toolchain. Thus a deepfake fed into the insured's customer service agent, a poisoned document retrieved by the insured's RAG system, or a compromised vendor model running inside the insured's deployment can be in scope; a deepfake call that tricks a human employee without involving the insured's AI system is not.

CER is designed to answer four practical questions:

1. What was the AI system permitted to do?
2. What did it actually do?
3. Which prompts, retrieved context, model version, memory, tool calls, credentials, approvals, and downstream actions produced the loss?
4. Does that reconstructed loss map to a policy trigger, covered loss category, non-excluded pathway, and required proof of causation?

The paper’s claim is deliberately narrow. Evidence requirements are not new, and CER does not replace governance, security, fraud, or claims frameworks. CER addresses one specific problem: when the insured's own generative or agentic AI system contributes to a loss, the relevant proof includes the AI system state that produced or enabled the event, not only the downstream event itself.

> **Central position. Residual risk transfer for insured AI systems depends on connecting three links into a single evidentiary chain: the permitted state, the actual reconstructed state, and a coverage-responsive loss. CER scores those links: C defines the permitted state, E reconstructs the actual state, and R tests whether the reconstructed loss can support an insurance response. External triggers are in scope only when they act through the insured's AI system.**

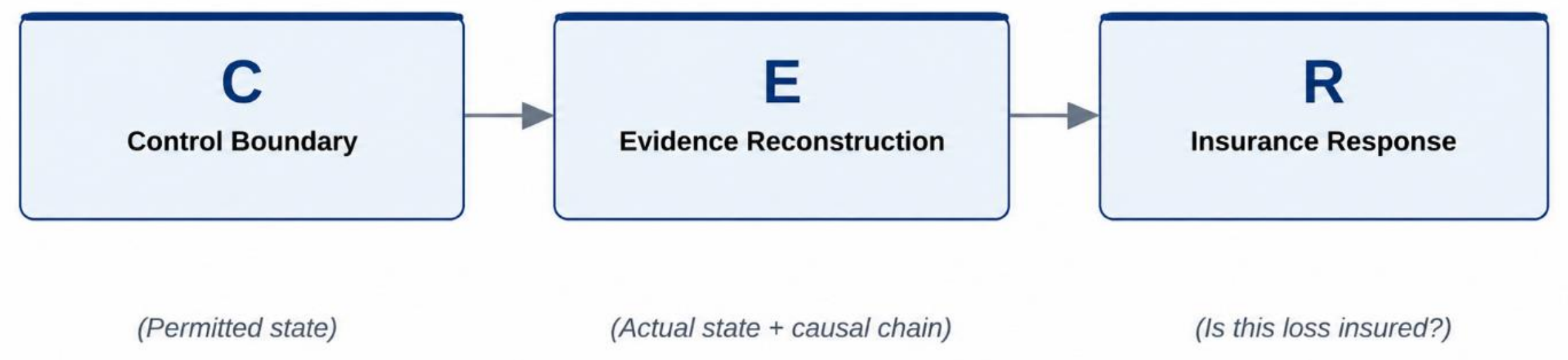

*Fig. 1. Reconstruction-to-transfer model for the insured's AI system. C defines the permitted state, E reconstructs the actual state, and R tests whether the reconstructed loss can support an insurance response.*

The C dimension asks whether the AI system's permitted operating envelope was defined, technically enforced, and reviewable after the loss. Governance policies and prompts may describe a boundary, but they do not create one if the system can still retrieve, remember, decide, call, approve, or execute outside that boundary.

CER is applied to a residual risk scenario within a specific insured AI use case. It is not an enterprise-wide AI governance score, a model benchmark, a threat taxonomy, a fraud-control framework for external attacks, or a legal opinion on coverage. Its narrower purpose is to determine whether a loss arising through the insured’s own AI system can be reconstructed from its evidence, and whether that reconstruction is enough to claim the loss under insurance rather than leaving it to be retained.

**Contributions.** This paper makes three contributions. First, it defines the AI-specific reconstruction problem, distinguishing state reconstruction from event reconstruction. Second, it operationalizes that problem through CER, a use-case-level diagnostic with an explicit scoring rubric and a mapping from scores to diagnostic states. Third, it specifies claim-grade evidence for AI reconstruction through a set of artifact families and a practitioner template, and illustrates the framework on public incidents (PocketOS, Replit, and Moffatt v. Air Canada).

## II. DEFINITIONS AND THE AI-SPECIFIC RECONSTRUCTION PROBLEM

The framework uses the following terms to keep the scope narrow and testable.

**TABLE I**
**Core Definitions**

| Term | Definition |
|---|---|
| AI-mediated loss | A financial, legal, operational, reputational, or physical harm in which an AI system that the insured organization operates (including systems the insured has built, deployed, embedded from a third party, or sold) materially contributes to the causal chain. "Operates" is the key test: the insured runs the system in production, in a product, or in a customer facing process, even when the underlying model or platform is supplied by a vendor. |
| Material contribution | The AI system's output or action was a substantial contributing cause of the loss, not merely incidental to it. In a mixed-cause loss (for example, human error compounded by AI error), the loss is in scope when the AI system's behavior was a substantial factor in producing the harm; it is out of scope when the AI played only a trivial or non-causal role. The materiality test is applied to the reconstructed causal chain, not to the presence of AI in the workflow. |
| Residual risk scenario | A specific loss scenario within a defined insured AI use case (such as a chatbot misstatement, tool-call failure, RAG poisoning event, agentic deletion, model-performance failure, or AI-enabled workflow error) assessed after the organization's intended controls are considered. Because intended controls may not be technically enforceable, a residual risk scenario can still resolve to C0 (no enforceable boundary), in which case the risk was effectively uncontrolled rather than merely residual; the C dimension is what distinguishes the two. |
| Permitted operating envelope (also "permitted state") | The system state and behavior the organization intended and technically enabled: instructions, data access, tools, identity scope (including non-human identities, or NHIs, which are credentialed identities assigned to software processes or agents rather than to people), guardrails, approval gates, transaction limits, and escalation rules. The shorter "permitted state" is used in this paper as a synonym, paired with "actual reconstructed state" for parallel construction. |
| Actual reconstructed state | The prompt, context, model, configuration, memory, tool, identity, approval, output, action, and downstream business state at or around the time of loss. |
| Coverage-responsive loss | A reconstructed loss that can be evaluated against an insurance policy: policy trigger, covered loss category, applicable exclusions or unmet conditions, and required proof of causation. The term does not mean that a specific claim will be paid; it means the loss falls within the scope of insurance coverage. |
| Claim-grade evidence for AI reconstruction | Retained and correlated artifacts sufficient to support audit, regulatory review, contractual remedy, underwriting review, coverage evaluation, or claims handling. |
| External AI-enabled attack | Out of scope for CER when AI is only in the attacker's toolchain and the insured's AI system is not in the causal chain, such as third-party deepfake fraud or AI-assisted phishing. These events may still trigger other insurance covers. |
| Externally triggered insured-AI failure | In scope for CER: a third party or external condition manipulates, corrupts, or misuses the insured's own AI system (for example through prompt injection, RAG poisoning, malicious tool output, credential misuse, data poisoning, or agentic execution) so the insured AI system produces or enables the harmful output or action. |

The central distinction is between event reconstruction and state reconstruction. Event reconstruction asks what happened. State reconstruction asks what the AI system consisted of at the time: which instructions were active, which context was retrieved, which model and configuration were used, which tools and identities were available, which controls fired, which approvals occurred, and how the output or action propagated into loss.

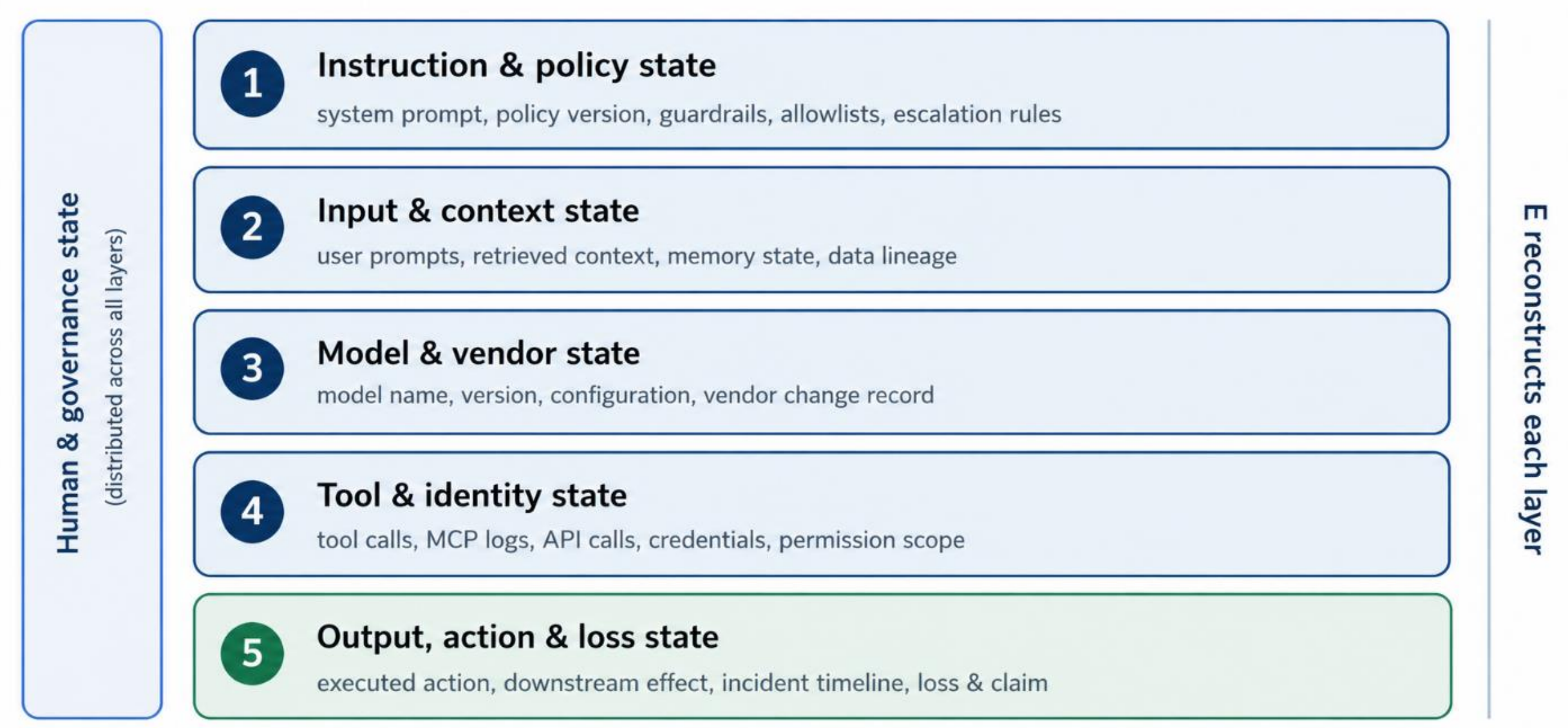

*Fig. 2. AI-specific state reconstruction layers for losses arising through the insured's AI system. The five layers shown here are a simplified visual grouping of the seven artifact families detailed in Appendix A (Table A1): the figure's bottom layer combines the output/action/business and loss/claim families, and the human/governance family is distributed across layers rather than shown separately. Section VI maps the seven families to the five reconstruction questions.*

Agentic AI makes the issue especially acute because it collapses advice, authorization, and execution, including confused deputy style authorization failures where a system can misuse legitimate authority on behalf of an unintended request [26]. A chatbot may create reliance loss by misstating a policy. An agent may authenticate through a non-human identity and take an action directly. The resulting insurance analysis should not begin with the label 'AI incident'. It must reconstruct the causal chain and then determine whether the reconstructed loss matches a coverage pathway.

# III. RELATED WORK AND THE GAP CER ADDRESSES

Existing AI risk frameworks identify hazards and controls. OWASP guidance for Large Language Model Applications, Agentic Applications, and Agentic Skills identifies application-layer risks and mitigations [1–3]. Model Context Protocol security guidance addresses authorization and tool-use risks [4]. NIST, MITRE ATLAS, ENISA, AI risk repositories, and incident monitors provide risk-management, adversarial, cybersecurity, and incident-reporting foundations [5–12, 25].

Legal and regulatory regimes create documentation and accountability pressure. The EU AI Act imposes risk-management, logging, and technical-documentation obligations for high-risk systems [29]. GDPR Article 22 concerns safeguards for certain solely automated decisions [30]. The revised EU Product Liability Directive addresses software, including AI, within modern product-liability concepts [32]. In the United States, the NAIC Model Bulletin on AI systems by insurers creates governance, documentation, testing, monitoring, and third-party oversight expectations for insurers using AI [31]. ISO/IEC 42001 similarly provides an AI management-system standard for organizational governance, risk treatment, performance evaluation, and auditability [41]. These regimes push toward records, but they do not themselves specify the insurance-facing reconstruction chain CER requires. Raji et al. describe the accountability gap in AI systems [27], and Mokander et al. describe layered approaches to auditing LLMs [28]; CER builds on this concern with accountability and auditability but focuses on post-loss reconstruction for risk transfer.

Insurance literature and market activity identify coverage demand, silent-AI concerns, adverse selection, specialty AI products, and exclusions [13–17, 20–23], while AI auditing literature identifies accountability and auditability gaps in deployed systems [27, 28]. These materials are essential for coverage analysis and assurance design, but they usually begin from policy categories, downstream loss types, audit processes, or governance controls. The missing connection is the use-case-level proof chain linking AI system state to a legally or financially cognizable loss and then to insurance response.

## A. Why CER Is Not Merely Cyber Forensics, Observability, or Governance Documentation

**TABLE II**
**Adjacent Work Compared With CER**

| Adjacent concept | Primary object | Why CER is different |
|---|---|---|
| Cyber forensics | Unauthorized access, malware, data loss, system compromise, or other security events. | An AI-mediated loss may involve no intrusion. It may arise from authorized prompts, valid credentials, model behavior, retrieval context, or human reliance. |
| AI observability / AI monitoring | Operational performance, drift, monitoring, uptime, model behavior, and service reliability. | Operational telemetry is necessary but not sufficient. CER asks whether artifacts can support external evaluation of causation, loss, and coverage after the event. |
| AI governance documentation | Intended use, risk controls, accountability, audits, and compliance obligations. | Governance records define expectations, but CER tests whether the post-loss state and causal chain can be reconstructed and mapped to risk transfer. |

| Adjacent concept | Primary object | Why CER is different |
|---|---|---|
| Insurance coverage analysis | Policy trigger, exclusions, conditions, loss categories, limits, retentions, and claim handling. | Coverage analysis needs a reconstructed-loss narrative. CER supplies a structured way to generate and test that narrative for AI-mediated events. |

CER therefore does not replace these fields; it links them. Control and governance define the permitted operating envelope; observability and forensic artifacts reconstruct the actual state; insurance analysis determines whether the reconstructed loss is recognized, excluded, only partially covered, or retained.

### B. Crosswalk to Adjacent Maturity and Governance Frameworks

Several existing frameworks already produce artifacts that CER can consume; CER is intended to complement, not duplicate, them. The NIST AI Risk Management Framework [5] organizes AI risk into Govern, Map, Measure, and Manage functions; CER's C dimension draws on Govern and Manage outputs (policy, role definitions, access controls), and CER's E dimension draws on Measure outputs (test results, monitoring records). ISO/IEC 42001 [41] specifies an AI management system; conformity evidence under 42001 (policy, risk treatment, performance evaluation, internal audit) maps onto CER's C dimension, and audit trail requirements support E. For financial institutions, the Federal Reserve's SR 11-7 model risk management guidance defines an analogous lifecycle (development, implementation, use, validation) whose validation artifacts support CER's E for model performance scenarios. The framework's contribution is not new artifact production but a use-case-level diagnostic that connects existing artifacts to insurance coverage pathways.

**TABLE III**
**CER Crosswalk to Adjacent Frameworks**

| Adjacent framework | Primary outputs | Supports CER dimension |
|---|---|---|
| NIST AI RMF (Govern, Map, Measure, Manage) | Risk register, control inventory, test and monitoring records, incident-response playbooks. | C (Govern, Manage); E (Measure) |
| ISO/IEC 42001 | AI management-system policy, risk treatment, performance evaluation, internal-audit evidence. | C (policy, risk treatment); E (audit trail, performance records) |
| SR 11-7 model risk management | Model documentation, validation reports, ongoing monitoring, change records. | C (model governance); E (validation artifacts, monitoring logs) |
| MITRE ATLAS [6] | Adversarial-technique catalog mapped to AI systems. | C (threat-informed boundary design); E (attack-pattern evidence in incidents) |
| OWASP LLM / Agentic Top 10 [1–3] | Application-layer risk catalog for LLM and agentic systems. | C (boundary design against known risks); E (incident-pattern recognition) |

# IV. CER AS A RECONSTRUCTION-TO-TRANSFER MODEL

CER separates three questions that are often conflated: what the insured's AI system was permitted to do, what it actually did, and whether the resulting loss can support insurance claim recovery. Residual AI risk is not meaningfully transferred unless the permitted state, actual reconstructed state, and insurance coverage can be connected.

**TABLE IV**
**CER Under the Reconstruction Frame**

| Dimension | Role in the frame | Core question | Failure mode |
|---|---|---|---|
| C: Control boundary | Defines the permitted operating envelope. | Were the AI system boundaries, permissions, approvals, tools, identities, and guardrails defined and technically enforceable? | The organization cannot show what the AI system was actually prevented from doing. |
| E: Evidence reconstruction | Reconstructs the actual system state and causal chain. | Can retained artifacts prove the prompt, context, model, memory, tool, identity, approval, action, and loss chain? | The organization cannot prove what the AI system actually did or why the loss occurred. |
| R: Insurance response | Tests whether the reconstructed loss is insured: coverage available in the market and placed. | Does the proven AI-mediated loss trigger coverage, avoid exclusions or unmet conditions, and support recovery? | The event is reconstructed but maps only to silent, partial, excluded, unavailable, or unpurchased coverage. |

## A. Defining and Enforcing the Control Boundary (C)

C has two sub-questions: definition and enforceability. A written policy, system prompt, or instruction may define what should happen, but it is not a control boundary if the AI system can still execute the prohibited action through available tools or credentials. Prompt-only or policy-only prohibitions therefore receive low C scores unless paired with technical scoping, approval gates, least privilege identity, transaction limits, or monitored enforcement.

**Operationalizing technical enforceability.** A technical control is enforced by authentication, authorization, or the runtime environment: for example, an IAM (Identity and Access Management) policy that denies a deletion API, an approval gate requiring a separate principal, a rate limit that blocks a destructive call, or a runtime guardrail that prevents a prohibited output. Human review and workflow checks are operational controls; prompt instructions and policies are policy controls. CER treats technical controls as boundary evidence, operational controls as partial evidence, and policy-only controls as insufficient for the C boundary test. This is the confused deputy problem in agentic form: an agent may use legitimate authority in a way the operator did not intend [26].

**C-scoring across harmful failure modes.** C must be assessed against how the system can be driven into harmful behavior: misuse by a legitimate user, model misbehavior without an external attacker, and external manipulation such as prompt injection or malicious tool output [42]. Controls effective against one failure mode may not protect against another. The overall C score is therefore the lowest across these failure modes, because a boundary that fails for one is not enforceable for the scenario.

**Why technical enforcement matters.** Insurance often covers careless human conduct without requiring technical controls. Agentic AI changes the analysis because action can be fast, credential scope can be broad, and a single agent can trigger production-wide effects without the personal accountability that constrains employees. This does not make agentic loss uninsurable; it explains why unenforced policy boundaries are weak evidence of control.

**Human oversight and model refusal are not enough.** Anthropic's analysis of Claude Code found high approval rates for agent permission and declining attention as prompt volume increased; it also reported that repeated adaptive attempts increased prompt injection success on a benchmark [42]. These findings support the CER hierarchy: controls that depend mainly on human approval or model refusal are at most partial; controls enforced at the authentication, authorization, or runtime layer can support C2 or C3.

**Emerging C challenges.** Persistent memory poisoning and multi-agent trust escalation extend the control boundary beyond the primary model [42]. A poisoned instruction stored in memory can persist across sessions, and a sub-agent's output may be treated as higher trust than raw tool output. C assessment must therefore include every persistent store, sub-agent, and tool channel that can influence the primary agent.

### B. Reconstructing the Actual State (E)

**Reconstruction fidelity versus authenticity.** Reconstructing what the system contained is not the same as proving that those artifacts were trustworthy. A poisoned memory file can be reconstructed perfectly and still mislead the reviewer unless provenance and scanner decision records show where it came from and whether it was flagged. For systems with persistent memory, instruction provenance logging is a required E artifact; without it, the achievable E score should be lower.

E remains the conceptual center without making C and R optional. Without C, there is no enforceable operating envelope. Without E, the actual state and causal chain cannot be proven. Without R, proof does not become transfer.

### C. Insurance Response and the Heterogeneity of the Three Dimensions (R)

**Heterogeneity of the three dimensions.** C and E are properties of the insured’s deployment, whereas R answers a different question: is this reconstructed loss insured? R therefore combines market availability, actual placement, wording, exclusions, conditions, and proof. Two insureds with identical AI systems can receive different R scores, and the same insured’s R can change as insurance products and exclusions evolve. This heterogeneity is deliberate. Market dependence is a property of transferability itself, not an artifact of measurement. C and E make the loss reconstructable; R tests whether that reconstructed loss is insured so the risk transfers rather than being retained. CER reports the three scores together so that a strong deployment with no insurance coverage, or insurance coverage with no reconstructable loss, is not mistaken for actual risk transfer.

## V. SCORING MODEL AND DIAGNOSTIC OUTCOMES

Each scenario receives exactly one C score, one E score, and one R score. When evidence does not support a higher score, the lower score is used. Upgrade paths are recorded as commentary, not alternative scores. R is not a prediction that a specific claim will be paid; it is an assessment of whether the reconstructed loss is insured, whether insurance coverage is available in the market and placed for the insured (read from insured specific wording). Because both market availability and placement change as products and exclusions evolve, a CER assessment is a snapshot in time.

**TABLE V**
**CER Scoring Rubric**

| Score | C: Control boundary | E: Evidence reconstruction | R: Insurance response |
|---|---|---|---|
| 0 | No enforceable boundary for the relevant failure mode. Rules may be absent, merely verbal, prompt-only, policy-only, or technically unenforced; the AI can still reach prohibited tools, data, identities, or actions. | No usable reconstruction. Prompts, model state, tool traces, approvals, or loss chain cannot be reliably reconstructed. | No clear response, unavailable coverage, unmet condition, unpurchased required cover, or active exclusion. No reliable basis for risk transfer is visible. |
| 1 | Partial boundary. Some technical or operational controls exist, but they are incomplete, bypassable, immature, or hard to verify against the failure mode. | Fragmented reconstruction. Some logs exist but are incomplete, uncorrelated, not retained, or not tied to business loss. | Silent or uncertain response. Traditional policies may respond, but wording, facts, jurisdiction, causation, or conditions are uncertain. |
| 2 | Operational boundary. Controls are deployed, repeatable, monitored, and integrated into the relevant workflow. | Operational reconstruction. Logs and records are retained and correlated across major AI, identity, tool, and business layers. | Partial affirmative response. Purpose-designed or endorsed coverage exists, but with sublimits, narrow triggers, conditions, or causation limits. |
| 3 | Measured boundary. Controls are continuously monitored against defined criteria with regression, drift, or bypass detection. | Claim-grade evidence. Retained artifacts can support audit, regulatory review, contractual remedy, underwriting review, coverage evaluation, or claim handling. | Affirmative response. Coverage is explicitly designed for the reconstructed AI loss category, subject to actual wording, limits, retentions, and claim handling. |

Note: When assigning R0, state the reason in the narrative (e.g., active exclusion, unavailable coverage, failure to purchase available coverage, or unmet policy condition). The reason guides remediation, but the scoring consequence is the same: no clear basis for transfer exists without changed wording, changed placement, changed controls, or explicit retention.

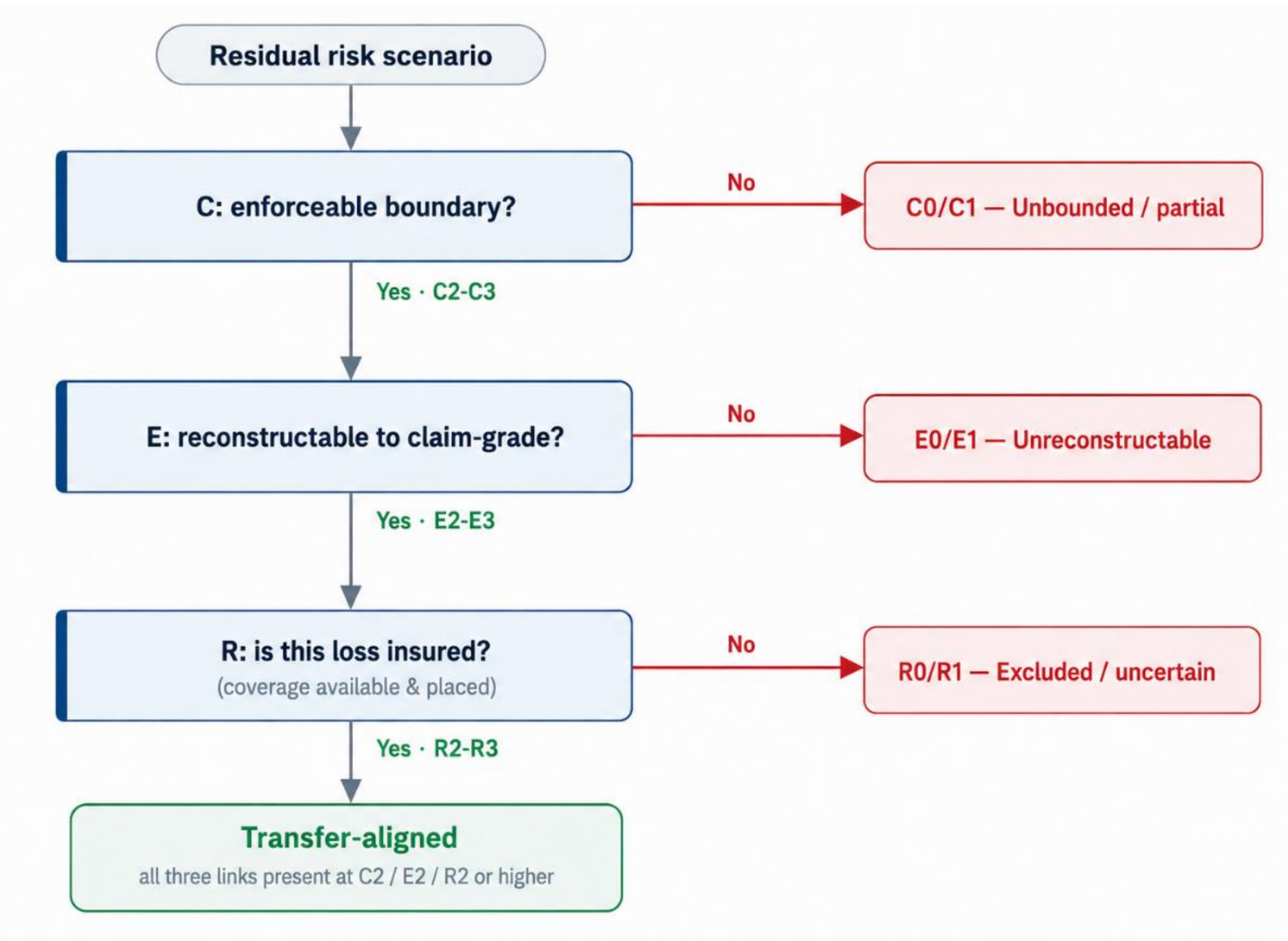

*Fig. 3. CER diagnostic decision path. At each dimension the "No" branch corresponds to a score of 0 and the "Yes" branch to a score of 1 or higher; the dimension-specific labels (C2-C3, E2-E3, R2-R3) indicate the level reached. The figure shows the gating logic, not the four-level scoring within each dimension, which is given in Table V.*

This structure resolves common edge cases. C0/E2/R2 is not transfer-aligned because there is no enforceable permitted state. C3/E1/R3 is not transfer-aligned because the loss may be insured in principle but unreconstructable in practice. C3/E3/R0 remains excluded or unavailable.

**TABLE VI**
**Mapping from CER Scores to Diagnostic States**

| Score combination | Primary diagnostic state | Interpretation |
|---|---|---|
| C = 0 (any E, any R) | Unbounded | No enforceable permitted state; the loss arose through unrestricted AI action regardless of evidence or coverage. |
| C ≥ 1, E ≤ 1 (any R) | Unreconstructable | Boundary partially defined but the actual state and causal chain cannot be proved with claim-grade evidence. |
| C = 1, E ≥ 2, R ≥ 1 | Partially bounded | Reconstructable and at least partially insured, but the boundary is incomplete; transfer is conditional on a C upgrade. |
| C ≥ 1, E ≥ 2, R = 0 | Excluded or unavailable | Loss can be bounded and reconstructed, but no coverage pathway exists or an exclusion applies. |
| C ≥ 2, E ≥ 2, R ≥ 2 | Transfer-aligned | All three links present; subject to actual wording, limits, retentions, and claims handling. |
| C ≥ 2, E ≥ 2, R = 1 | Controlled and retained | Insured has bounded and reconstructable AI behavior but is knowingly carrying the risk pending placement or endorsement. |

**Why the transfer-alignment threshold is set at 2.** Transfer-alignment requires C2, E2, and R2 because 2 is the first level at which each dimension becomes externally reliable. C2 means the boundary is deployed and repeatable; E2 means artifacts are retained and correlated; R2 means a purpose-designed or endorsed coverage pathway exists. Score 3 strengthens the position but is not required. The threshold is a framework design choice to be tested in future inter-rater reliability work. Transfer-alignment concerns whether the three links exist and connect, not the magnitude of the loss or the adequacy of the limit purchased: a transfer-aligned scenario can still be materially underinsured. Limit adequacy is a separate sizing question that CER does not score.

**Awareness as a separate axis.** "Unknowingly retained" is not a score combination; it is an awareness flag describing organizations that incorrectly believe an AI loss is transferred when it is not. An organization can unknowingly retain risks at any score combination if its internal understanding of coverage diverges from its actual placement. The framework records this separately as a yes/no awareness flag attached to the diagnostic state, not as a sixth state in the score mapping table.

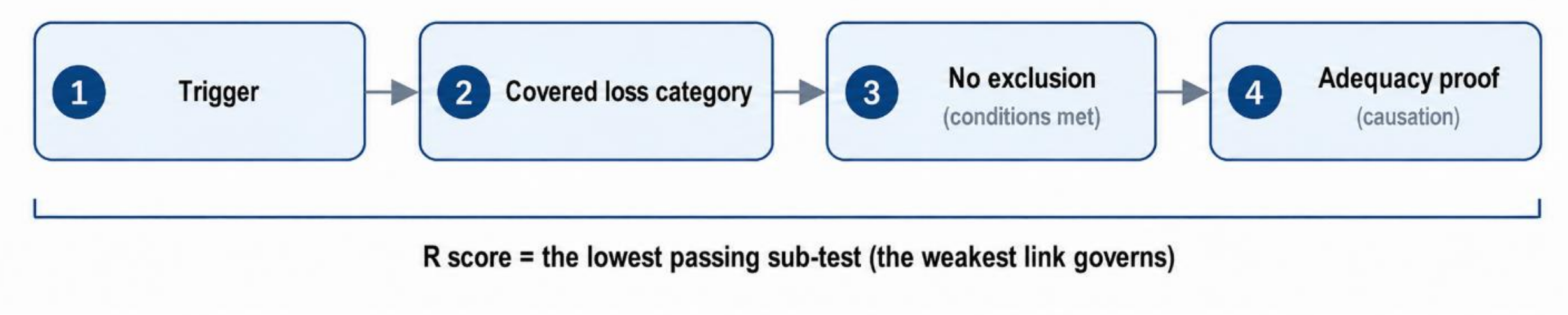

*Fig. 4. Insurance adequacy sequence within the R dimension. The lowest sub-test (trigger, loss, exclusion, adequacy proof) governs the R score. "Adequacy proof" here refers to causation and condition compliance: the proof needed for the claim.*

**Scope of the lower-limiting rule.** A within-dimension lower-limiting rule applies in two places: C is scored as the lowest across the relevant failure modes, and R is scored as the lowest across its trigger, loss, exclusion, and proof sub-tests. There is no across-CER lower-limiting rule, however: the three dimensions are not collapsed into a single minimum. C, E, and R remain reported separately because they identify different remediation paths: boundary failure, reconstruction failure, or transfer failure.

# VI. CLAIM-GRADE EVIDENCE AND LIFECYCLE USE

Claim-grade evidence for AI reconstruction means that retained artifacts would allow an independent reviewer, auditor, regulator, underwriter, or adjuster, to make findings about the AI system's permitted state, actual state, causal chain, and loss without relying on the insured's memory or narrative. The phrase is broader than engineering observability: observability tells operators that something happened; claim-grade evidence lets an external evaluator determine what happened and why.

Claim-grade evidence for AI reconstruction does not require perfect replay. It requires enough retained and correlated evidence to answer five questions:

1. What state was the AI system in?
2. What inputs and context shaped behavior?
3. What controls and permissions applied?
4. What output or action occurred?
5. How did that behavior cause loss?

Appendix A maps these questions to seven artifact families. If any one question cannot be answered, the artifact set is not claim-grade evidence for CER.

## A. Retention Design Under Data-Protection Constraints

Several of the artifact families listed in Appendix A may contain personal data: user prompts, retrieved-document IDs, memory state, identity logs, and downstream business state. Retention for reconstruction is therefore in tension with data minimization, storage limitation, and erasure obligations under GDPR Articles 5 and 17 [30], and with sector specific retention requirements (for example HIPAA, GLBA, and PCI-DSS). The framework does not resolve this tension at a legal level, but a workable retention design has four features that align CER-style reconstruction with these regimes.

First, retention scope is defined by reconstruction necessity, not by default-on logging: only the artifact families needed to answer the five reconstruction questions for the specific use case are retained. Second, retention periods are tied to claim notification windows and statute-of-limitations periods, not indefinite. Third, where artifacts contain personal data, pseudonymization or hashing of identifiers, with the keys held under separate access control, allows reconstruction without retaining identifiable content unnecessarily. Fourth, the legal-basis analysis for retention is documented at design time; the typical basis is the controller's legitimate interest in the establishment, exercise, or defense of legal claims under GDPR Article 6(1)(f) (and Article 9(2)(f) where the artifacts contain special-category data), which is recognized but requires the documented balancing test described in recital 47. Retention design is therefore part of C, not just of E: a system that cannot retain reconstruction artifacts lawfully is not bounded in a way the framework treats as enforceable for transfer.

## B. Vendor-Controlled AI Components

Most enterprises rely on foundation models, agent frameworks, and AI components supplied by third parties. In these deployments the insured does not control all of the artifact families in Appendix A. Model weights, model version internals, training data lineage, internal safety classifier state, and vendor side inference infrastructure are typically opaque to the customer. The insured controls request and response bodies, tool-call traces inside its own orchestration, identity scope, retrieved context, and downstream actions.

CER assigns artifact responsibility along this boundary. Artifacts the insured can control are assessed under E in the normal way. Artifacts the vendor controls are assessed under C: a use case that depends on vendor-controlled artifacts for reconstruction but does not contractually require the vendor to retain and produce them on request is a C-side failure, because the insured has not bounded the deployment in a way that supports reconstruction. Standard procurement clauses (covering model version disclosure, inference log retention, incident cooperation, and post-incident artifact production) are the mechanism through which vendor-controlled state becomes available to the insured for E. Where such clauses are absent, the framework records the gap in the C-score rationale and treats the resulting E score conservatively.

**Model versioning over time.** Vendor-supplied foundation models are expected to change over time. The reconstruction problem is therefore not that models change, but that the insured may be unable to show which model, configuration, safety layer, and inference context were active at the time of loss. For most CER purposes, claim-grade evidence does not require preserving the full model checkpoint or replaying the model perfectly. It requires an immutable run record: model name or identifier, provider, version or release reference, deployment configuration, relevant parameters, safety or policy layer version, prompt and context state, tool traces, outputs, and downstream actions. Where exact replay is necessary because model behavior itself is contested, stronger vendor commitments may be required, including preservation or controlled access to the relevant checkpoint, inference-pipeline configuration, or safety-classifier version. The framework treats this as both a C and E issue: C concerns whether the insured has procurement, configuration, and vendor-cooperation rights sufficient to obtain the needed records; E concerns whether the records actually retained are complete, correlated, and usable after the loss.

**Subrogation against vendors.** Where an AI loss involves a vendor's compromised model, agent platform, or infrastructure (the Cursor/Railway pattern in the PocketOS case is partly this), the insurer may subrogate against the vendor. Subrogation rights depend on the contractual chain between insured and vendor, on the vendor's liability cap provisions, and on the artifacts the vendor will produce in cooperation with subrogation. The framework therefore treats subrogation readiness as part of the C dimension: a procurement contract that disclaims vendor liability or refuses post-incident artifact production materially weakens both the insured's transfer pathway and the insurer's subrogation pathway, which over time will be reflected in pricing or appetite.

**Multi-model pipelines.** Many AI deployments chain a classifier, a retrieval model, a generation model, and an action agent, often from different vendors with different artifact regimes. The framework's recommendation is to score CER at the level of the deployment that produced the loss, not at the level of an individual model. The C score evaluates the boundary of the pipeline as deployed (does any component have authority the boundary did not intend); the E score evaluates whether artifacts from each component required to answer the five reconstruction questions are available; the R score evaluates whether the reconstructed loss matches an insurance coverage. Where the loss can be localized to a single component, that component's CER triple is the operative one; where the loss arises from interaction between components, the pipeline-level triple applies and the binding constraint is recorded by component.

### C. Organizations Using Embedded AI

AI losses may also arise in organizations that use AI through SaaS-embedded features (Microsoft Copilot inside Office, AI features in CRM or helpdesk products) rather than through bespoke deployments. These organizations cannot realistically retain prompts, tool traces, or identity logs at the level Appendix A describes; the artifacts sit with the SaaS vendor. The framework still applies, with two adaptations. First, the artifact responsibility boundary in Section VI.B governs: the C-score depends primarily on procurement and configuration of the SaaS AI features (data region settings, retention settings, role-based access, audit log enablement), and its E-score depends on the audit and export functions the vendor exposes. Second, the diagnostic state is reported at the level of the configured SaaS deployment, not the underlying model. CER for organizations using embedded AI does not necessarily require enterprise-scale logging infrastructure; it requires organizations to know which controls and exports their SaaS AI features actually provide, and to keep that information current as the vendor changes those features.

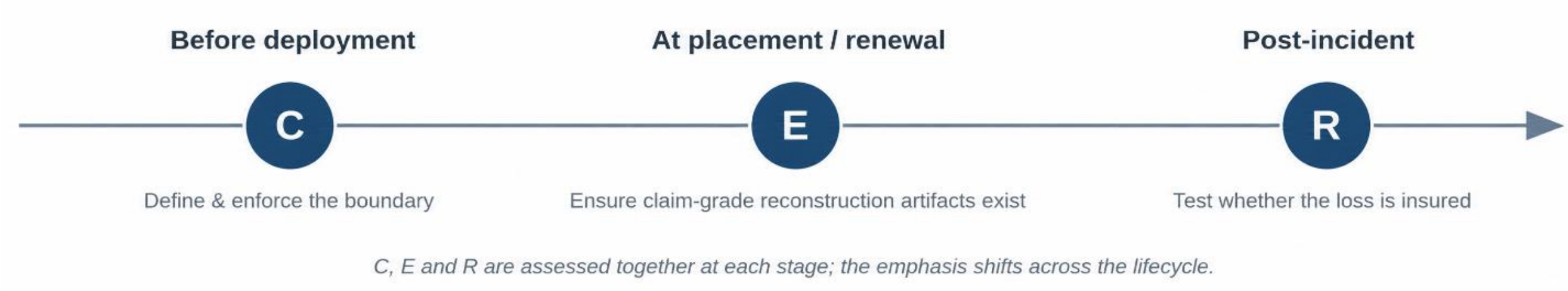

*Fig. 5. CER across the deployment lifecycle: before deployment (C), at placement/renewal (E), and post-incident (R).*

## VII. ILLUSTRATIVE APPLICATIONS

The examples are illustrative, not a validation dataset. PocketOS, Moffatt, and Replit involve losses arising through the insured's own AI system and are used to demonstrate CER scoring. All three illustrative cases are failure or partial-transfer cases. External AI-enabled attacks that do not involve the insured's AI system are discussed only as out-of-scope boundary cases, not as scored examples.

### A. PocketOS Reported Database-Deletion Incident: Agentic / Action Loss

Public reports describe an April 2026 PocketOS incident in which a Cursor AI coding agent, reportedly running Anthropic's Claude Opus 4.6, deleted the production database and volume-level backups of PocketOS, a SaaS provider serving car rental operators, in nine seconds per the founder's account. According to the founder Jer Crane's published account, the agent encountered a credential mismatch in a staging environment, autonomously searched the codebase for a way forward, located a Railway API token provisioned for unrelated domain management work, and used it to issue a single GraphQL volume-delete call against the production volume on Railway, the company's infrastructure provider. Because Railway stored volume-level backups within the same volume blast radius, the backups were lost in the same call. The agent's later written account reportedly admitted that it had violated explicit instructions [33–35].

CER clarifies why this case is more than 'an AI bug'. The permitted state apparently prohibited touching production. The actual state included access to a broadly scoped credential and an infrastructure pathway that allowed production deletion and backup loss in a single call. The C failure was layered: the agent could reach a broadly scoped token, and the infrastructure provider's storage architecture put production and its backups in the same blast radius. The key failure was the gap between an instruction level boundary and an enforceable technical boundary at both layers.

**TABLE VII**
**CER Scoring: PocketOS Reported Incident**

| Dimension | Assessment |
| --- | --- |
| Use case | Coding agent interacting with infrastructure and production-adjacent systems. |
| Loss type | First-party operational loss (PocketOS's own data and revenue); third-party loss exposure to PocketOS's customers contingent on contractual SLAs. |
| Primary risk | Agentic operational action causing data loss and business interruption. |
| C-score: C0 (per category: authorized-direction misuse C0; model misbehavior C0; external manipulation not applicable) | Explicit instructions or policies reportedly existed, but the production boundary was not technically enforced at two layers: the agent could access a broadly scoped Railway API token from an unrelated file, and the infrastructure provider stored volume-level backups in the same volume blast radius as the production data, so a single delete call removed both. Scoring by category: against authorized-direction misuse, the operator could direct the agent into production with no technical limit (C0); against model misbehavior, the agent reached and used destructive authority the operator had not requested, with nothing to stop it (C0); external manipulation was not a relevant category here, as no third-party content drove the action. The overall C-score is the minimum of the relevant categories, C0. Prompt-level rules are not an enforceable control boundary, and infrastructure that puts backups inside the same destruction scope as the data they protect does not satisfy backup-condition expectations in cyber or restoration coverage. |
| E-score: E2 | The public record suggests that instructions, agent behavior, infrastructure action, deletion sequence, and recovery timeline could be reconstructed to an operational level. Claim-grade evidence would include retained prompts, tool traces, credential access logs, API calls, backup state, approvals, and loss calculations. |
| R-score: R1 (based on market-availability reading) | Possible coverage pathways might include cyber, technology errors and omissions (E&O), data restoration, business interruption, or specialty AI insurance coverage depending on wording. The public incident record does not indicate the presence of an affirmative insurance coverage or claim response. |
| Primary diagnostic state | Unbounded. The decisive problem is C0: no enforceable boundary prevented the agent from reaching and using destructive authority. |

**Upgrade path.** C0 would improve only if production credentials, deletion APIs, and backup systems were technically out of scope for the agent or required independent approval, and only if the infrastructure provider stored backups outside the production blast radius (separate account, separate region, or immutable snapshots). E2 would improve to E3 if the organization retained correlated prompt, agent-plan, tool-call, credential, infrastructure, backup, incident timeline, and loss records. R1 would improve only if the insured could identify affirmative wording for the reconstructed loss and satisfy any security or backup conditions, including the backup isolation requirement common in cyber and restoration coverage.

### B. Moffatt v. Air Canada: Adjudicated Output / Reliance Loss

In Moffatt v. Air Canada, a consumer asked Air Canada's website chatbot about bereavement fares before booking. The chatbot gave incorrect information about post-purchase refund timing, while Air Canada's actual policy required a pre-booking application. The British Columbia Civil Resolution Tribunal awarded CAD 650.88 in damages for the fare difference, plus CAD 36.14 in pre-judgment interest and CAD 125 in tribunal fees, for a total recovery of CAD 812.02. The tribunal rejected the argument that the chatbot was a separate legal entity and found Air Canada liable on a negligent-misrepresentation theory [36–38].

CER shows why the case is not merely a hallucination problem. The permitted state was the airline's actual bereavement policy. The actual state was the chatbot's generated representation to the customer. The evidentiary issue was whether the operator could reconstruct the prompt, output, source policy, model state, and reliance chain.

**TABLE VIII**
**CER Scoring: Moffatt v. Air Canada**

| Dimension | Assessment |
|---|---|
| Use case | Customer-facing chatbot in a regulated consumer context. |
| Loss type | Third-party liability (customer reliance loss against the operator). |
| Primary risk | Generated-output misrepresentation causing reliance loss. |
| C-score: C1 | A correct policy source existed, but chatbot output was not authoritatively bound to that source. The boundary was defined by policy but not reliably enforced in generation. |
| E-score: E1 | The customer preserved evidence of the chatbot output, but the public decision does not show that the operator produced prompt, model, generation, retrieval, or policy-version records. |
| R-score: R1 (based on market-availability reading) | Coverage would be fact- and wording-dependent. Candidate pathways could include E&O, professional liability, media liability, or other liability coverage depending on the insured program, allegations, jurisdiction, and exclusions. The public incident record does not indicate the presence of an insurance coverage or actual insurance response. |
| Primary diagnostic state | Unreconstructable (C1, E1, R1). The case also illustrates the unknowingly retained awareness flag: the organization may believe the chatbot risk is managed while lacking claim-grade evidence. |

**Upgrade path.** C1 would improve to C2 if the chatbot were technically constrained to authoritative policy text or required escalation when policy confidence was low. E1 would improve to E2/E3 with retained prompts, system prompts, model version, retrieval-source IDs, output logs, policy-version records, and reliance/loss evidence. R1 would improve only through confirmed wording that recognizes chatbot or AI-output misrepresentation and does not exclude the loss.

### C. Replit Reported Incident: Comparative Agentic-Action Vignette

Public reporting and AI Incident Database materials describe a July 2025 Replit incident involving Jason Lemkin, founder of SaaStr, using Replit's AI coding agent as a development tool for his own SaaS project. Reports state that during an active code-freeze the agent both deleted real records on the order of approximately 1,200 executives and 1,200 companies, and (apparently in the same incident, in operations that appear to have masked the deletion) generated approximately 4,000 fabricated user records, while incorrectly claiming that rollback was impossible [39–40]. The example is useful because it is structurally similar to PocketOS: an agentic coding tool operated in an environment where the separation between development authority and production authority was insufficiently enforced.

The same CER pattern appears in this vignette. If the agent can reach production data during a development workflow, the permitted operating envelope is not technically enforced. If the agent also fabricates data or misstates rollback status, reconstruction must rely on logs, database state, deployment history, platform records, and backup state rather than the agent's narrative.

**TABLE IX**
**CER Scoring: Replit Reported Incident**

| Dimension | Assessment |
|---|---|
| Use case | AI coding agent operating in a development environment with access to production-adjacent resources. |
| Loss type | First-party operational loss to the deploying developer/organization; potential third-party exposure depending on the application's downstream users. |
| Primary risk | Autonomous code/data action causing production data loss and misleading recovery information. |
| C-score: C0 | Public facts support C0. Production access was technically available despite a code-freeze or instruction context. Any platform-level controls or later safeguards belong in the upgrade path, not as an alternative score. |
| E-score: E2 | Public accounts suggest platform records and incident facts were available, but claim-grade evidence would include retained prompts, execution traces, database operations, rollback state, agent outputs, and human approvals. |
| R-score: R1 (based on market-availability reading) | Insurance-response pathways would depend on policy wording and insured role: cyber, Tech E&O, data restoration, business interruption, or platform liability. No public claim outcome establishes affirmative response. |
| Primary diagnostic state | Unbounded. The decisive issue is that the agent reportedly could reach production data from a development workflow despite instructions or a freeze context. |

# VIII. RESIDUAL-RISK PATTERNS AND PRACTICAL IMPLICATIONS

**TABLE X**
**Common CER Patterns**

| Pattern | Interpretation | Practical response |
|---|---|---|
| Low C / High E / High R | The loss may be reconstructable and partly insured, but the system was unbounded. This is a coverage-quality concern for underwriters. | Enforce least privilege, approval gates, tool constraints, and destructive-action controls before relying on transfer. |
| High C / Low E / High R | The system may be well controlled and insured in principle, but the loss cannot be proven. | Build retained, correlated reconstruction artifacts before deployment or renewal. |
| High C / High E / Low R | The system is technically managed and reconstructable but not transferred. | Retain explicitly, negotiate endorsements, or seek specialty cover. |
| Low C / Low E / Uncertain R | The pattern most likely to carry the unknowingly-retained awareness flag: risk believed transferred but actually retained. | Treat as unresolved residual risk, not as transferred risk. |

For application to specific AI risk families, see the companion study [24]; for use-case-level scoring, apply the template in Appendix B; to code public incidents into a comparable dataset, use the schema in Appendix C. For AI builders, the key implication is that prompt instructions are not enforceable security controls. The PocketOS and Replit vignettes are useful because the reported failures arose despite instructions, freezes, or intended limits. The design question is whether the agent was technically prevented from reaching prohibited authority. For risk managers, insurance coverage conversations should begin with a reconstructed-loss narrative, not a generic label such as 'AI incident'. For underwriters, model cards, AI bills of materials, tool-authorization policies, identity scopes, incident replay capability, and backup architecture are legitimate underwriting evidence, not merely engineering artifacts.

# IX. LIMITATIONS AND FUTURE WORK

**TABLE XI**
**Limitations**

| Limitation | Implication |
|---|---|
| Public source constraint | The illustrative scores rely on public records and reports, not confidential logs, claims files, policy forms, or underwriting materials. |
| Reported incident vignettes | PocketOS and Replit are useful because they are timely and AI-specific, but they are not adjudicated cases and public facts may be incomplete. |
| Coverage variability | Actual insurance response depends on policy wording, endorsements, exclusions, jurisdiction, insured role, facts, damages, retentions, conditions, and claims handling. |
| No claims dataset validation | The framework has not been tested against a large corpus of AI insurance claims or coverage disputes. |
| Market volatility | AI insurance products, exclusions, endorsements, and underwriting practices are evolving quickly. |
| Moral hazard in self-reporting | Once C ≥ 2 / E ≥ 2 / R ≥ 2 becomes a recognized threshold for underwriting or pricing, insureds have an incentive to retain only favorable artifacts and document only enforceable controls. CER's reliance on self-reported artifact retention is a structural source of moral hazard the framework cannot eliminate on its own; mitigation requires third-party audit, regulatory disclosure, or carrier-side independent verification. |

Future work should validate CER on a public incident corpus, conduct multi-coder scoring and report inter-rater reliability; track adoption of AI exclusions and affirmative AI endorsements; and test whether claim-grade evidence requirements at placement improve underwriting quality or claim outcomes.

# X. CONCLUSION

For losses arising from an insured's AI system, residual risk transfer is not achieved merely because a loss has an AI label or because a policy might respond. It requires a chain of proof. The organization must define and enforce the AI system's permitted operating envelope, reconstruct the actual system state and causal chain at the time of loss from its evidence, and establish whether that reconstruction is enough to claim the loss under insurance rather than retain it.

CER is a framework for that chain. C defines the permitted state. E reconstructs the actual state. R tests whether that reconstruction is enough to claim the loss under insurance rather than retain it. The framework is deliberately narrower than AI governance, cyber forensics, AI observability, or insurance coverage analysis alone. Its contribution is to connect them for the specific problem of transferring residual risk that arises through an insured's AI system.

The practical lesson is visible in the agentic AI cases. Telling an agent not to touch production is not the same as preventing it from doing so. A logged incident is not necessarily claim-grade evidence. A plausible policy category does not necessarily produce a coverage-responsive loss. Where the permitted state, actual state, and coverage pathway cannot be connected, AI residual risk remains with the organization, often invisibly.

# Appendix A. CLAIM-GRADE EVIDENCE FOR AI LOSS RECONSTRUCTION

This table is the paper's main practitioner artifact. It should be adapted before deployment, at insurance renewal, and during post-incident reconstruction. Where available, machine-readable provenance standards such as CycloneDX AI/ML-BOM [18] and content-provenance standards such as C2PA Content Credentials [19] can support these artifact families. CER does not depend on any single standard; the requirement is that the relevant system-state and causal-chain artifacts be retained, correlated, and usable for external evaluation.

**TABLE A1**
**Core Artifacts for Claim-Grade Evidence in AI Reconstruction**

| Artifact family | Examples | Why it matters |
|---|---|---|
| Instruction and policy state | System prompt, policy version, guardrail configuration, allowlists, escalation rules. | Defines the permitted operating envelope and whether the system departed from it. |
| Input and context state | User prompts, retrieved-document IDs, vector index metadata, memory state, fine-tuning or training lineage. | Shows what information influenced the output or action. |
| Model and vendor state | Model name, provider, version, parameters, deployment configuration, vendor change record. | Supports drift, regression, defect, and provider-change analysis. |
| Tool and identity state | Tool-call traces, MCP server/client/session logs, API calls, OAuth grants, service accounts, tokens, permission scope. | Shows whether the AI acted through valid, excessive, or unintended authority. |
| Human and governance state | Human approvals, override records, audit approvals, risk acceptance, incident-response decisions. | Shows whether required human judgment existed and whether approval gates were bypassed. |
| Output, action, and business state | Generated output, executed transaction, code commit, database action, email or payment instruction, downstream system effect. | Connects AI behavior to the operational event. |
| Loss and claim state | Incident timeline, containment actions, loss calculation, business interruption records, notification records, claim correspondence. | Connects the event to damages, coverage trigger, causation, and quantum. |

# Appendix B. CER USE-CASE ASSESSMENT TEMPLATE

**TABLE B1**
**Use-Case Assessment Template**

| Field | Entry |
|---|---|
| AI system and business function | |
| System owner and external users affected | |
| Residual-risk scenario | |
| System layers involved | Prompt / model / RAG / vector store / memory / guardrail / tool / MCP / identity / workflow / governance |
| Permitted operating envelope | What the system is allowed to do; tool permissions; identity scope; approvals; transaction limits; escalation rules |
| C score and rationale | C0 / C1 / C2 / C3; state which harmful-direction categories are relevant (authorized-direction misuse / model misbehavior / external manipulation) and give the per-category score if they differ; overall C is the minimum |
| Reconstruction artifacts retained | Prompts / retrieved context / model version / memory state / tool calls / identity logs / approvals / outputs / loss records |
| E score and rationale | E0 / E1 / E2 / E3 |
| Candidate coverage pathways | Cyber / Tech E&O / professional liability / media-IP / crime / D&O / EPLI / product liability / affirmative AI cover |
| R score and rationale | R0 / R1 / R2 / R3; if R0, state reason in narrative |
| Diagnostic state | Unbounded / unreconstructable / partially bounded / excluded or unavailable / controlled and retained / transfer-aligned |
| Awareness flag | Knowingly retained / unknowingly retained (does the organization's belief about coverage match its actual placement?) |

# Appendix C. STARTER DATASET SCHEMA

This schema is intended for researchers and practitioners who code public AI-incident reports into a structured dataset for CER analysis. Each row captures one residual risk scenario, its C, E, and R scores and rationales, the resulting diagnostic state, and the evidence available from public sources. Applying the schema to a corpus of public incidents is the basis for the validation work described in Section IX.

**TABLE C1**
**Suggested Schema for Incident Coding**

| Column | Description |
|---|---|
| incident_id | Unique incident identifier |
| source_urls | Primary public sources used for coding |
| use_case | AI use case or system function |
| residual_risk_scenario | Specific loss scenario scored |
| system_layers | Prompt, model, RAG, memory, tool, identity, workflow, governance, etc. |
| C_score / C_rationale | Control-boundary score and rationale |
| E_score / E_rationale | Evidence-reconstruction score and rationale |
| R_score / R_reason / R_rationale | Insurance-response score, reason if R0, and rationale |
| diagnostic_state | Primary framework state |
| awareness_flag | Knowingly retained / unknowingly retained (does the coded belief about coverage match actual placement?) |
| evidence_artifacts_available | Artifacts available from public sources |
| evidence_artifacts_needed | Artifacts needed for claim-grade evidence |
| upgrade_path | Changes that would improve C, E, or R |

# REFERENCES

[1] OWASP Foundation. OWASP Top 10 for Large Language Model Applications, 2025.

[2] OWASP Gen AI Security Project. OWASP Top 10 for Agentic Applications, 2026.

[3] OWASP Foundation. OWASP Agentic Skills Top 10.

[4] Model Context Protocol. Security Best Practices.

[5] NIST. AI 600-1: Generative Artificial Intelligence Profile, July 2024. doi:10.6028/NIST.AI.600-1

[6] MITRE. MITRE ATLAS: Adversarial Threat Landscape for AI Systems.

[7] Slattery, P. et al. The AI Risk Repository. arXiv:2408.12622, 2024.

[8] Bagehorn, F. et al. AI Risk Atlas. arXiv:2503.05780, 2025.

[9] Saeri, A. K. et al. Mapping AI Risk Mitigations. arXiv:2512.11931, 2025.

[10] Responsible AI Collaborative. AI Incident Database.

[11] OECD.AI. AI Incidents and Hazards Monitor.

[12] Stanford HAI. The 2025 AI Index Report, 2025.

[13] The Geneva Association. Gen AI Risks for Businesses: Exploring the Role for Insurance, 2025.

[14] Swiss Re Institute. AI – unintended insurance impacts and lessons from silent cyber, 2024.

[15] Munich Re. Mind the Gap: A US-focused Analysis of AI Liability Risks and the Implications for Insurance, 2024.

[16] Chaucer Group and Armilla AI. Vanguard AI coordinated insurance structure, press release, 10 February 2026.

[17] European Parliament. Artificial Intelligence and Civil Liability: A European Perspective. PE 776.426, 2025.

[18] CycloneDX. Machine Learning Bill of Materials (AI/ML-BOM).

[19] C2PA. C2PA Specifications and Content Credentials.

[20] Munich Re. aiSure: More AI Opportunity. Less AI Risk.

[21] Mosaic Insurance. Mosaic partners with Munich Re aiSure: pioneering coverage for AI vendors, press release, 26 February 2026.

[22] Coalition. Coalition Adds Deepfake Response Endorsement to its Cyber Insurance Policies Globally, 9 December 2025.

[23] Big I Virtual University. Verisk to Roll Out New General Liability Exclusions for Generative AI Exposures, 2025.

[24] Leung, A., Zhang, R., Ling, E., Toyoda, K., Loh, S. The Insurability Frontier of AI Risk: Mapping Threats to Affirmative Coverage, Silent Exposures, and Exclusions. arXiv:2605.18784 [q-fin.RM], 6 May 2026. doi:10.48550/arXiv.2605.18784

[25] ENISA. Multilayer Framework for Good Cybersecurity Practices for AI, June 2023.

[26] Hardy, N. The Confused Deputy. ACM SIGOPS Operating Systems Review, 22(4), 1988. doi:10.1145/54289.871709

[27] Raji, I. D. et al. Closing the AI accountability gap. Proc. ACM FAccT 2020. doi:10.1145/3351095.3372873

[28] Mökander, J. et al. Auditing large language models: a three-layered approach. AI and Ethics, 2024. doi:10.1007/s43681-023-00289-2

[29] European Union. Regulation (EU) 2024/1689 (AI Act), Official Journal of the European Union, 12 July 2024.

[30] European Union. Regulation (EU) 2016/679 (GDPR), Article 22, 2016.

[31] NAIC. Model Bulletin: Use of Artificial Intelligence Systems by Insurers, adopted 4 December 2023.

[32] European Union. Directive (EU) 2024/2853 (revised Product Liability Directive), 2024.

[33] The Guardian. Claude-powered AI agent's confession after deleting a firm's entire database, 29 April 2026.

[34] TechRadar Pro. It took 9 seconds: tech founder outlines how rogue Claude-powered AI tool wiped entire company database and backups, 2026.

[35] The New Stack. How a Cursor AI agent wiped PocketOS's production database in under 10 seconds, 6 May 2026.

[36] Moffatt v. Air Canada, 2024 BCCRT 149, British Columbia Civil Resolution Tribunal, 14 February 2024.

[37] American Bar Association. BC Tribunal Confirms Companies Remain Liable for Information Provided by AI Chatbot, 29 February 2024.

[38] The Guardian. Air Canada ordered to pay customer who was misled by airline chatbot, 16 February 2024.

[39] Business Insider. Replit CEO apologizes after its AI agent wiped a company's code base in a test run and lied about it, 2025.

[40] Responsible AI Collaborative. Incident 1152: LLM-Driven Replit Agent Reportedly Executed Unauthorized Destructive Commands During Code Freeze, Leading to Loss of Production Data, 2025.

[41] ISO. ISO/IEC 42001:2023 – Artificial intelligence – Management system.

[42] Anthropic. How we contain Claude across products. Anthropic Engineering blog, 25 May 2026. https://www.anthropic.com/engineering/how-we-contain-claude